%% file: iclr2025_conference.tex
\documentclass{article} 
\usepackage{iclr2025_conference,times}

\input{math_commands.tex}

\usepackage{hyperref}
\usepackage{url}
\usepackage{booktabs}
\usepackage{graphicx}
\usepackage[utf8]{inputenc}

\title{MatWheel: Addressing Data Scarcity in Materials Science Through Synthetic Data}

\author{Wentao Li\thanks{These authors contribute equally}\ \ \  Yizhe Chen\footnotemark[1] \ \ \  Jiangjie Qiu  \\
Department of Chemical Engineering\\
Tsinghua University\\
\texttt{\{liwt24,chenyizh23,qjj21\}@mails.tsinghua.edu.cn} \\
\And
Xiaonan Wang\thanks{Corresponding author} \\
Department of Chemical Engineering \\
Tsinghua University \\
\texttt{\{wangxiaonan\}@tsinghua.edu.cn}
}

\iclrfinalcopy 
\begin{document}

\maketitle
\begin{abstract}
Data scarcity and the high cost of annotation have long been persistent challenges in the field of materials science. Inspired by its potential in other fields like computer vision, we propose the MatWheel framework, which train the material property prediction model using the synthetic data generated by the conditional generative model. We explore two scenarios: fully-supervised and semi-supervised learning. Using CGCNN for property prediction and Con-CDVAE as the conditional generative model, experiments on two data-scarce material property datasets from Matminer database are conducted. Results show that synthetic data has potential in extreme data-scarce scenarios, achieving performance close to or exceeding that of real samples in all two tasks. We also find that pseudo-labels have little impact on generated data quality. Future work will integrate advanced models and optimize generation conditions to boost the effectiveness of the materials data flywheel.
\end{abstract}

\section{Introduction}

The use of synthetic data is particularly significant in fields like computer vision, where augmenting data can help improve the generalization of deep learning models. For example, synthetic images generated through Generative Adversarial Networks (GANs) \citet{goodfellow2014generative} or Variational Autoencoders (VAEs) \citet{kingma2013auto} have been used successfully to train models for tasks such as image classification and segmentation \citet{zhu2017unpaired} \citet{shorten2019survey}. These synthetic data augmentations improve model robustness by exposing the model to a more diverse set of examples. However, as the quality of generated data improves and as models train on synthetic data, issues related to bias and distribution shift can arise. This challenge of balancing data enrichment and model bias is not unique to vision or nature language processing (NLP) but extends into material science.

Materials science, similarly, has seen a growing interest in using generative models to overcome the lack of experimental data. These models, particularly conditional generative models, aim to generate molecular structures or material properties conditioned on predefined characteristics such as bandgap, formation energy, or stability. These models can generate a large number of materials that meet specific property requirements, thus accelerating the discovery of novel materials. For example, MatterGen \citet{zeni2025generative} and Con-CDVAE \citet{ye2024cdvae} have demonstrated the ability to generate materials that satisfy specific properties, showing the potential of generative models in materials design. However, there is limited research on the impact of synthetic data in materials science, and no experimental insights have been provided on whether it can establish a materials data flywheel.

Our work addresses this gap by investigating the use of synthetic materials data for improving property prediction models. We explore whether synthetic data generated from a conditionally trained generative model can enhance the predictive performance of property models when incorporated into training. This research is inspired by similar approaches in other fields, where iterative training on synthetic data has shown potential to improve model accuracy \citet{you2024diffusion}, but with the caution that the quality and distribution of synthetic data must be carefully managed. In materials science, where data collection is expensive and time-consuming, the ability to effectively generate and use synthetic data for improving predictive models could significantly accelerate the pace of materials discovery.

\begin{figure}[h]
\begin{center}
\label{Figure1}
\includegraphics[width=\columnwidth]{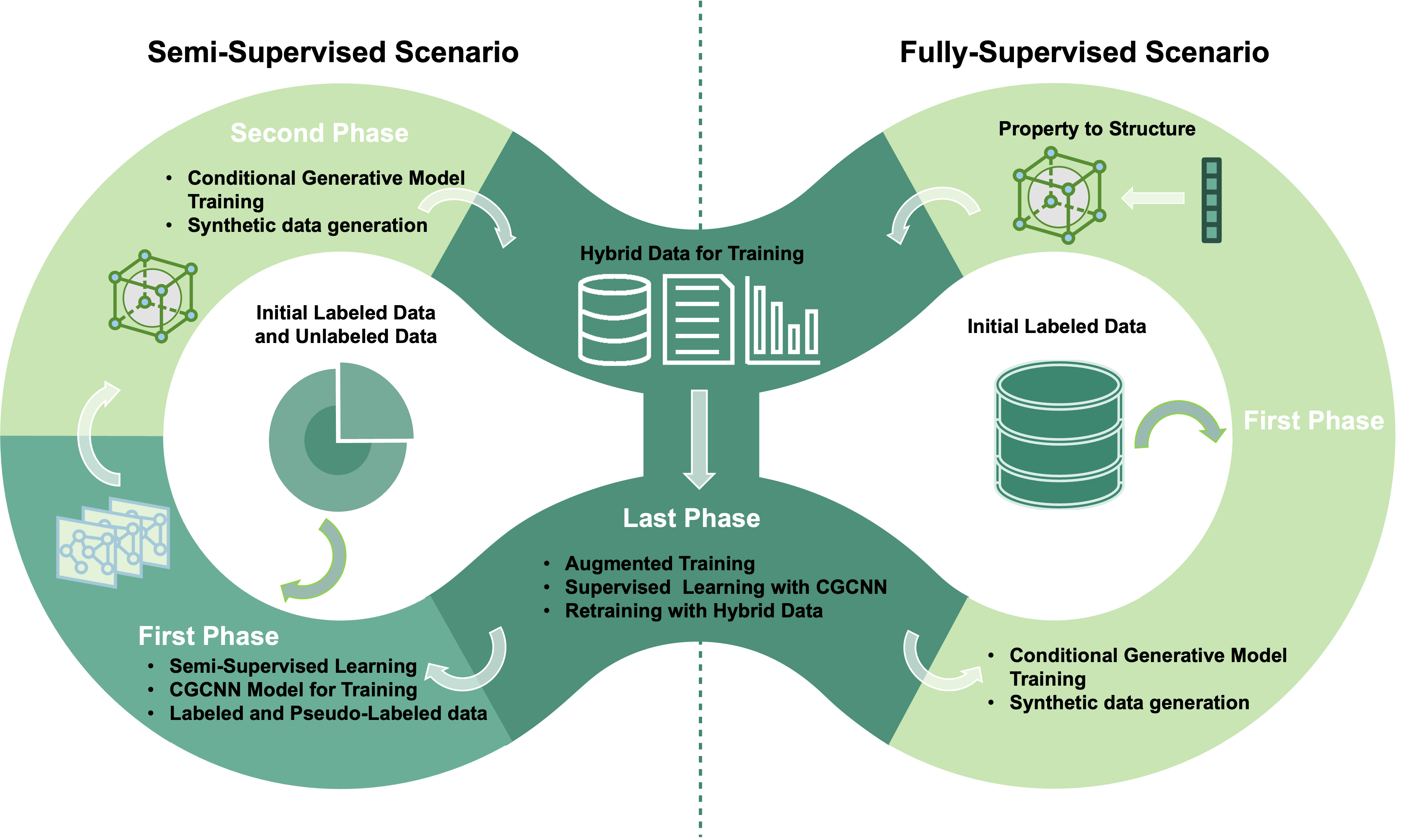}
\end{center}
\caption{MatWheel overall framework. In the semi-supervised case we divided into three stages of training and inference, in the full sample case we divided into two stages of training and inference}
\end{figure}

We introduce the concept of a "data flywheel" in materials science and propose a framework called MatWheel, where synthetic data generated by a conditional generative model is used to improve both the generative model and the property prediction model, as shown in Figure~\ref{Figure1}. Through this iterative process, we aim to enhance the performance of predictive models by continuously integrating new synthetic materials data into the training pipeline. This approach not only leverages the power of generative models but also ensures that the iterative generation of data leads to meaningful improvements in model accuracy and robustness.

\section{Methodology}
Our research framework considers two distinct scenarios: a fully supervised learning scenario and a semi-supervised learning scenario.
	
1.	Fully Supervised Learning Scenario: This scenario is designed to evaluate whether synthetic data can enhance the performance of a predictive model trained on the full dataset. Here, the conditional generative model is trained using all available training samples. During inference, synthetic material structures are sampled using scalar properties as conditions, generating a synthetic dataset. The predictive model is then trained on a combination of real and synthetic training data.
	
2.	Semi-Supervised Learning Scenario: This scenario aims to assess whether the predictive model and the conditional generative model can evolve within the proposed data flywheel framework. Initially, the predictive model is trained on only k\% (In the specific experiment, we selected 10\%) of the training samples, and pseudo-labels are generated for all training samples through inference. The conditional generative model is subsequently trained using the entire training set, including pseudo-labeled data, and is used to sample an expanded synthetic dataset. Finally, the predictive model is retrained using a combination of the synthetic dataset and the 10\% real-labeled training data. The newly predicted pseudo - label data is likely to be more accurate. Moreover, we can expand the unlabeled dataset by introducing external databases or through unconditional generation, thus achieving the effect of an iterative cycle. 

We follow the dataset processing methodology of \citet{chang2022towards} and split the dataset into training (70\%), validation (15\%), and test (15\%) sets. For the semi-supervised learning scenario, the training set is further divided in a 1:9 ratio, resulting in a final split of 7:63:15:15, where the 63\% portion represents the pseudo-labeled training data generated by the predictive model.We selected two material property prediction tasks characterized by data scarcity, with each dataset containing no more than 1,000 samples. These datasets represent the common form and quantity of material data. The data were sourced from the Matminer \citet{ward2018matminer} database. The dataset information is presented in Table~\ref{Table1}.

\begin{table}[t]
\caption{Statistical Information of Data-Scarce Material Property Datasets}
\label{Table1}
\vspace{1em} 
\centering
    \begin{tabular}{ccccccc}
        \hline
        \bfseries Datasets & \bfseries Total number & \bfseries \begin{tabular}[c]{@{}c@{}}Maximum Number \\ of Atoms\end{tabular} & \bfseries Property range \\
        \hline
        Jarvis2d exfoliation & 636 & 35 & (0.03, 1604.04) \\
        MP poly total & 1056 & 20 & (2.08, 277.78) \\
        \hline
    \end{tabular}
\end{table}

We selected CGCNN \citet{xie2018crystal} as the property prediction model. CGCNN leverages the architecture of graph convolutional neural networks, enabling effective processing of atomic spatial relationships and features within crystal structures. For the conditional generative model, we adopted Con-CDVAE, an improved version of CDVAE \citet{xie2021crystal}. Con-CDVAE incorporates scalar properties as input and applies diffusion to atomic counts, atomic species, spatial coordinates, and lattice vectors, ultimately generating target materials that closely align with the specified property values. During model training and validation, we followed the hyperparameters used in both original studies for training and inference.

For numerical sampling in the conditional generative model, we perform kernel density estimation (KDE) on the discrete distribution of the training data. In the fully supervised scenario, the KDE is constructed based on the 70\% real training data, while in the semi-supervised scenario, it is based on the combined 7\% real training data and 63\% pseudo-labeled data. Sampling is then conducted from the estimated KDE distribution to generate conditional inputs for the generative model.

\section{Results \& Discussions}

We conducted five independent random runs on two data-scarce material property prediction datasets. For each scenario, we computed the mean and standard deviation of the five results. The final outcomes are presented in Table~\ref{Table2}.

F represents training the predictive model using the full training dataset. $G_F$ denotes training the conditional generative model on the full training dataset, generating 1,000 synthetic samples, which are then used to train the predictive model. F+$G_F$ refers to training the predictive model on a combination of the full real training dataset and the 1,000 synthetic samples. The same methodology applies to the semi-supervised scenario. $G_S$ denotes training the conditional generative model on the pseudo-labels training dataset, while the remaining procedure follows the same approach as in the fully supervised setting.

\begin{table}[htbp]
\caption{Impact of Synthetic Data on the Predictive Model in Two Scenarios}
\label{Table2}
\vspace{1em} 
    \centering
    \begin{tabular}{lcccccc}
        \toprule
        & \multicolumn{3}{c}{\textbf{Fully-Supervised}} & \multicolumn{3}{c}{\textbf{Semi-Supervised}} \\
        \cmidrule(r){2-4}\cmidrule(l){5-7}
        \textbf{Datasets} & \multicolumn{1}{c}{\textbf{F}} & \multicolumn{1}{c}{$\textbf{G}_{\textbf{F}}$} & \multicolumn{1}{c}{$\textbf{F}\textbf{+}\textbf{G}_{\textbf{F}}$} & \multicolumn{1}{c}{\textbf{S}} & \multicolumn{1}{c}{$\textbf{G}_{\textbf{S}}$} & \multicolumn{1}{c}{$\textbf{S}\textbf{+}\textbf{G}_{\textbf{S}}$} \\
        \midrule
        Jarvis2d \\ exfoliation & $62.01_{\text{12.14}}$ & $64.52_{\text{12.65}}$ & $\bm{57.49_{\textbf{13.51}}}$ & $64.03_{\text{11.88}}$ & $64.51_{\text{11.84}}$ & $\bm{63.57_{\textbf{13.43}}}$ \\
        MP poly \\ total & $\bm{6.33_{\textbf{1.44}}}$ & $8.13_{\text{1.52}}$ & $7.21_{\text{1.30}}$ & $8.08_{\text{1.53}}$ & $8.09_{\text{1.47}}$ & $\bm{8.04_{\textbf{1.35}}}$ \\
        \bottomrule
    \end{tabular}
\end{table}

We observe that in the fully supervised scenario, training solely with synthetic data yields the poorest performance. Combining synthetic data with real data for training shows a notable improvement only on the Jarvis2d exfoliation dataset, whereas in the MP poly total dataset, this approach underperforms compared to training exclusively on real data.

In the semi-supervised scenario, mixing synthetic data with real data achieves the best results on the Jarvis2d exfoliation and MP poly total datasets. We also observe that the performance of $G_F$ is very similar to  $G_S$, and in some cases, $G_S$ even outperforms $G_F$. This suggests that conditional generative model does not necessarily require real structure-property pairs, as pseudo-labels do not significantly affect the quality of the generated data.

This observation further indicates that the inherent bias of the conditional generative model is greater than that of the predictive model, rendering the impact of pseudo-labels negligible. Consequently, this highlights the necessity of defining evaluation metrics to assess the effectiveness of the conditional generative model, establishing a benchmarking framework, and reducing the discrepancy between the property distribution of generated structures and that of real materials.

\section{Conclusion}

We conducted experiments on two data-scarce material property datasets and successfully implemented the fundamental framework of the materials data flywheel. The experimental results partially demonstrate the potential of synthetic data in empowering the materials flywheel, while also highlighting the current limitations of synthetic data generation.

Through our analysis, we identified key issues in the generated data and their possible causes. Moving forward, we plan to integrate the more advanced model (MatterGen) and refine the design of generation conditions and sampling distributions. This will enable synergistic co-evolution between the generative model and the predictive model, ultimately enhancing the effectiveness of the materials data flywheel.

\bibliography{iclr2025_conference}
\bibliographystyle{iclr2025_conference}

\end{document}

%% file: math_commands.tex
\usepackage{amsmath,amsfonts,bm}

\def\eqref#1{equation~\ref{#1}}

\def\1{\bm{1}}

\DeclareMathAlphabet{\mathsfit}{\encodingdefault}{\sfdefault}{m}{sl}
\SetMathAlphabet{\mathsfit}{bold}{\encodingdefault}{\sfdefault}{bx}{n}

